\documentclass{article}

\usepackage[preprint]{neurips_2022}

\usepackage[utf8]{inputenc} % allow utf-8 input
\usepackage[T1]{fontenc}    % use 8-bit T1 fonts
\usepackage{hyperref}       % hyperlinks
\usepackage{url}            % simple URL typesetting
\usepackage{booktabs}       % professional-quality tables
\usepackage{amsfonts}       % blackboard math symbols
\usepackage{nicefrac}       % compact symbols for 1/2, etc.
\usepackage{microtype}      % microtypography
\usepackage{xcolor}         % colors
\usepackage{subcaption,graphicx,listings,algorithm,amsmath}

\title{Mathematically Modeling the Lexicon Entropy of Emergent Language}

\author{Brendon Boldt, David Mortensen \\
Language Technologies Institute \\
Carnegie Mellon University\\
Pittsburgh, PA 15213, USA \\
\texttt{\{bboldt,dmortens\}@cs.cmu.edu} \\
}

\newcommand\theProcess{\textsc{FiLex}}
\newcommand\Nodyn{\textsc{NoDyn}}
\newcommand\Sig{\textsc{Sig}}
\newcommand\Nav{\textsc{Nav}}
\newcommand\Recon{\textsc{Recon}}

\begin{document}

\maketitle

\begin{abstract}
We formulate a stochastic process, \theProcess{}, as a mathematical model of lexicon entropy in deep learning-based emergent language systems.
Defining a model mathematically allows it to generate clear predictions which can be directly and decisively tested.
We empirically verify across four different environments that \theProcess{} predicts the correct correlation between hyperparameters (training steps, lexicon size, learning rate, rollout buffer size, and Gumbel-Softmax temperature) and the emergent language's entropy in $20$ out of $20$ environment-hyperparameter combinations.
Furthermore, our experiments reveal that different environments show diverse relationships between their hyperparameters and entropy which demonstrates the need for a model which can make well-defined predictions at a precise level of granularity.
\end{abstract}

\section{Introduction}

The methods of deep learning-based emergent language provide a uniquely powerful way to study the nature of language and language change.
In addressing these topics, some papers hypothesize general principles describing emergent language.
For example, \citet{resnick_capacity_2020} hypothesize a predictable relationship exists between compositionality and neural network capacity,
    and \citet{kharitonov2020entropy} hypothesize a general entropy minimization pressure in deep learning-based emergent language.
In many cases, these hypotheses are derived from intuitions and stated in natural language; this can lead to ambiguous interpretation, inadequate experiments, and \emph{ad hoc} explanations.
To this end, we study a general principle of emergent language by proposing a mathematical model which generates a testable hypothesis which can be directly evaluated through the empirical studies, akin to what we find prototypically in natural science.

We formulate a stochastic process, \theProcess{}, as a mathematical model of lexicon entropy in deep learning-based emergent language systems\footnotemark{} (ELS).
\footnotetext{\emph{Emergent language system} or ELS refers to the combination of agents (neural networks), the environment, and the training procedure used as part of an emergent language experiment.}
% Defining a model mathematically allows it to generate clear predictions which can be directly and decisively tested. \cmg{(re)move preceding sentence?}
We empirically verify across four different environments that \theProcess{} predicts the correct correlation between hyperparameters (training steps, lexicon size, learning rate, rollout buffer size, and Gumbel-Softmax temperature) and the emergent language's entropy in $20$ out of $20$ environment-hyperparameter combinations.

There are three primary reasons for using an explicitly defined model for studying a topic like emergent language: clarity, testability, and extensibility.
A mathematical model yields a \emph{clear}, unambiguous interpretation since its components have precise meanings; this is especially important when conveying such concepts in writing.
It is easier to \emph{test} a model than a hypothesis articulated in natural language because the model yields clear predictions which can be shown to be accurate or inaccurate; as a result, models can also be directly compared to one another.
Our experiments reveal that different environments show diverse relationships between their hyperparameters and entropy which demonstrates the need for such clarity in making well-defined predictions at a precise level of granularity.
Finally, mathematical models hypothesize a \emph{mechanism} for an observed effect and not simply the effect itself (with a possibly \emph{ad hoc} explanation).
This is what facilitates their \emph{extensibility} since a multitude of hypotheses can be derived from these mechanisms; furthermore, this ``mechanical'' nature allows future work to build directly on top of the model.

As mathematical models are seldom used to their full potential in studying emergent language, this paper is meant to serve as a reference and starting point for entire methodology of developing and testing such models.
We articulate our contributions as follows:
\begin{itemize}
    \item Defining a mathematical model of lexicon entropy in emergent language systems which we demonstrate to be accurate in predicting hyperparameter-entropy correlations.
    \item Presenting a case study of defining and empirically evaluating a mathematical model in emergent language.
    \item Provide a direct, intuitive comparison of the effects of hyperparameters on lexicon entropy across different environments.
\end{itemize}

We briefly discuss related work in Section~\ref{sec:related}.
In Section~\ref{sec:methods}, we introduce the mathematical model, \theProcess{}, as well as the ELSs.
Empirical evaluation is presented in Section~\ref{sec:experiments} and discussed in Section~\ref{sec:discussion}, concluding with Section~\ref{sec:conclusion}.
Code is available at \url{https://github.com/brendon-boldt/filex-emergent-language}.

\section{Related work}\label{sec:related}
For a survey of deep learning-based emergent language work, please see \citet{lazaridou2020survey}.
Contemporary deep learning-based emergent language research often aims at establishing and refining general principles about emergent language.
In large part, these principles can be expressed as relationships between certain characteristics of the environment or agents (e.g., model capacity \citep{resnick_capacity_2020}, population size \citep{rita2022on}) and properties of the emergent language (e.g., compositionality \citep{resnick_capacity_2020,rodriguezluna2020pressures}, entropy \citep{kharitonov2020entropy,chaabouni2021color,rita2022on}, and generalizability \citep{chaabouni2020compositionality,guo_expressivity_2021,slowik_exploring_2020}).
Some of these works \citep{kharitonov2020entropy,khomtchouk2018integral,resnick_capacity_2020} make use of mathematical models to describe parts of the hypotheses and/or experiments, but these fall short of establishing a clear model which generates a testable hypothesis which is then evaluated through the empirical studies.

Pre-deep learning emergent language research frequently relied on mathematical models \citep{skyrms2010signals,Kirby_Tamariz_Cornish_Smith_2015,Brighton2005LanguageAA}, but such models played a different role.
Whereas these models were meant to account for some property of language observed in human language, the model presented in this paper is accounting for emergent language directly (and human language only indirectly).
Thus, this paper presents a (mathematical) model of a (computational) model which, in the future, will be used to more directly study human language.

% On a methodological level, this paper is similar to \citet{resnick_capacity_2020} which also develops a theoretical model for an emergent language phenomenon.
% Namely, their model predicts the compositionality of an emergent language dependent on the capacity of the neural-network agents.
% Analogously, out model predicts the entropy of the emergent language dependent on learning rate, replay buffer size, bottleneck size, or training steps.

\section{Methods}\label{sec:methods}

\subsection{Model}
\theProcess{} (``fixed lexicon stochastic process'') is a mathematical model developed from the Chinese restaurant process \citep{blei2007crp,aldous1985exchangeability}, a stochastic process where each element in the sequence is a stochastic distribution over the positive integers (i.e., a distribution over distributions).
The analogy for the Chinese restaurant process is a restaurant with tables indexed by the natural numbers; as each customer walks in, they sit at a random table with a probability proportional to the number of people already at that table.
The key property here is that the process is \emph{self-reinforcing}; tables with many people are likely to get even more.
By analogy to language, the more a word is used the more likely it is to continue to be used.
For example, speakers may develop a cognitive preference for it, or it gets passed along to subsequent generations as a higher rate \citep{Francis2021}.
% Furthermore, self-reinforcing processes based on the Chinese restaurant process have been used to model the power-law distributions found in natural language \citep{goldwater2011pitmanyor}.

\paragraph{Formulation}
\newcommand\bs{\boldsymbol}
\theProcess{} is defined as a sequence of stochastic vectors indexed by $N\in\mathbb N^+$ given by:
\begin{align}
    \theProcess{}(\alpha, \beta, S, N)
        % &= \frac{\bs{w}^{(N)}}{\sum_{i=1}^S w^{(N)}_i} \\
        &= \frac{\bs{w}^{(N)}}{\|\bs w^{(N)}\|_1} \label{eq:filex} \\
    \bs w^{(n+1)} &= \bs w^{(n)} + \alpha\frac{\bs{x}^{(n)}}\beta \label{eq:update} \\
    \bs x^{(n)} &\sim \text{Multi}\!\left(\beta, \frac{\bs w^{(n)}}{\|\bs w^{(n)}\|_1}\right) \label{eq:sample} \\
    % w^{(1)}_i &= \frac1S \quad\text{for}\quad 1 \le i \le S \label{eq:init}
    % \bs w^{(1)} &= \frac1S \cdot \bs 1_S \label{eq:init} \\
    % \bs w^{(1)} &= \frac1S \cdot \begin{bmatrix}1&1&\cdots&1\end{bmatrix}^\top \in \mathbb R^S
    \bs w^{(1)} &= \frac1S \cdot (1,1,\ldots,1) \in \mathbb R^S
\end{align}
where
    $\bs w^{(n)}$ is a vector of weights,
    $\alpha \in \mathbb R_{>0}$ controls the weight update magnitude,
    $\beta \in \mathbb N^+$ controls the variance of the updates,
    $S \in \mathbb N^+$ is the size of the weight vector (i.e., lexicon),
    and $\text{Multi}(k, \bs p)$ is a $k$-trial multinomial distribution with probabilities $\bs p \in \mathbb R^S$.
The pseudocode describing \theProcess{} is given in Algorithm~\ref{alg:the-process}.
Conceptually, the process starts with an $S$-element array of weights initialized to $1/S$.
At each iteration we draw from a $\beta$-trial multinomial distribution parameterized by the normalized weights.\footnote{The $\beta$-trial multinomial sample is written as $\beta$ i.i.d.\@ samples from a categorical distribution to draw parallels to PPO in Algorithm~\ref{alg:ppo}.}
This multinomial sample is multiplied by $\alpha/\beta$ and added to the weights so that the update magnitude is $\alpha$.
This proceeds $N$ times.
Since the sequence elements are the \emph{normalized} weights, the elements are themselves probability distributions; thus, \theProcess{} is technically a sequence of distributions over distributions.

\lstset{
    escapeinside={(*@}{@*)},
    language=Python,
    emph={for,int,float,in,return,range},
    emphstyle={\color{blue}\ttfamily},
    basicstyle={\ttfamily\footnotesize},
    commentstyle={\color{gray}\footnotesize},
    numbers=left,
    xleftmargin=2.0em,
}

\begin{algorithm}[t]
        \begin{lstlisting}
alpha: float > 0
beta: int > 0
N: int > 0
S: int > 0

weights = array(size=S)
weights.fill(1 / S)
for _ in range(N):
  W += sample_multinomial(W / sum(W), beta) / beta
  w_copy = weights.copy()
  for _ in range(beta):  # equivalent to normalized multinomial
    i = sample_categorical(w_copy / sum(w_copy)) (*@\label{lst:sample}@*)
    weights[i] += alpha / beta
return weights / sum(weights)
        \end{lstlisting}
    \caption{\theProcess{} pseudocode}\label{alg:the-process}
\end{algorithm}

The two key differences between \theProcess{} and the Chinese restaurant process are the hyperparameters $S$ and $\beta$.\footnote{Note that $\alpha$ in \theProcess{} is actually equivalent to the \emph{inverse} of $\alpha$ in the Chinese restaurant process.}
\theProcess{} has a fixed number of parameters so as to match the fact that the agents in the ELS have a fixed-size bottleneck layer, that is, a fixed lexicon.
% \drm{Isn't this definitionally incompatible with being a stochastic process? I feel like we need a different terms.}
Secondly, $\beta$ is introduced to modulate the smoothness of parameter updates.
It is closely connected to the fact that certain RL algorithms like PPO accumulate a buffer of data points from the environment with the same parameters before performing gradient descent.

\subsection{Environments}
To evaluate \theProcess{}, we use four different reinforcement learning environments in our experiments.
These are inhabited by two deep learning-based agents: (1) a sender agent which receives an observation and produces a message and (2) a receiver agent which receives a message (and possibly additional observation) and takes an action.
The agent architecture and optimization are detailed Section~\ref{sec:agents}.

\paragraph{\Nodyn}
The ``no dynamics'' environment is a proof-of-concept environment which is not intended to be realistic but rather to match as closely as possible the simplifying assumptions which \theProcess{} makes while keeping the same neural architecture in the environments below.
As the name suggests, the primary simplification in this environment is that there are trivial dynamics, that is, every episode immediately ends with reward of $1$ no matter what the sender or receiver do.
The sender input and receiver output are identical to those of \Nav{}, defined below.
Just as \theProcess{} assumes that every instance of word use is reinforced, this process reinforces every message which the sender produces.

\paragraph{\Recon}
The reconstruction game \citep{chaabouni2020compositionality}, in the general case, mimics a discrete autoencoder: the input value is translated into a discrete message by the sender, and the receiver tries to output the original input based on the message.
For a given episode, the sender observes $x \sim \mathcal U(-1, 1)$ and produces a message; the receiver's action is a real number $\hat x$, yielding a reward $(x - \hat x)^2$.

\paragraph{\Sig} The signaling game environment comes from \citet{lewis1970convention} and has been frequently used in the literature \citep{lazaridou2016multiagent,bouchacourt2018agents}.
In this setup, the data is partitioned into a fixed number of discrete classes.
The sender observes a datum from one of the classes and produces a message; the receivers observes this message, the sender's datum, and data points from other classes (i.e., ``distractors'').
The reward for the environment is $1$ if the receiver correctly identifies the sender's datum among the distractors and $0$ otherwise.

To eliminate the potential confounding factors from using natural inputs (e.g., image embeddings \citep{lazaridou2016multiagent}), we use a synthetic dataset.
For an $n$-dimensional signaling game, we have $2^n$ classes.
Each class is represented by an isotropic multivariate normal distribution with mean $(\mu_1, \mu_2, \dots, \mu_n)$ where $\mu_i\in\{-3,3\}$.
Observations of a given are samples of its corresponding distribution.
For example, in the $2$-dimensional game, the $4$ classes would be represented by the distributions:
    $\mathcal N((-3, -3), I_2)$,
    $\mathcal N((3, -3), I_2)$,
    $\mathcal N((-3, 3), I_2)$,
    and $\mathcal N((3, 3), I_2)$
    (we use a $5$-dimensional signaling game for our experiments with $32$ classes).
The motivation for this setup is minimal need for feature extraction while still using real-valued, stochastic inputs.

\paragraph{\Nav}
For a multi-step environment, we use a $2$-dimensional, obstacle-free navigation task.
The sender agent observes the $(x,y)$ position of a receiver and produces a message; the receiver moves by producing an $(x,y)$ vector.
For a given episode, the receiver is initialized uniformly at random within a circle and must navigate towards a smaller circular goal region at the center.
The agents are rewarded for both reaching the goal and for moving towards the center.
An illustration is provided in Appendix~\ref{sec:illustration}.
% There are no obstacles or walls in the environment.
The receiver's location and action are continuous variables.
% If the receiver does not reach the goal region within a certain number of steps, the episode ends with no reward given.

\subsection{Agents}\label{sec:agents}

\paragraph{Architecture}
Our architecture comprises two agents, conceptually speaking, but in practice, they are a single neural network.
The sender and receiver are randomly initialized at the start of training, are trained together, and are tested together.
The sender itself is a $2$-layer perceptron with tanh activations.
The sender's input is environment-dependent.
The output of the second layer is passed to a Gumbel-Softmax bottleneck layer \citep{maddison2017concrete,jang2017categorical} which enables learning a discrete, one-hot representation.%
\footnote{Using a Gumbel-Softmax bottleneck layer allows for end-to-end backpropagation, making optimization faster and more consistent than using a backpropagation-free method like REINFORCE \citep{kharitonov2020entropy,williams1992simple}. Nevertheless, future work may want to use REINFORCE for its more realistic assumptions about communication.}
% as an information bottleneck \citep{tishby2000information}.
The activations of this layer can be thought of as the words forming the lexicon of the emergent language.
Messages consist only of a single one-hot vector (word) passed from sender to receiver.
At evaluation time, the bottleneck layer functions deterministically as an argmax layer, emitting one-hot vectors.
The receiver is a $1$-layer perceptron which takes the output of the Gumbel-Softmax layer as input.
The receiver's output is environment-dependent.
An illustration and precise specification are provided in Appendices~\ref{sec:illustration} and~\ref{sec:hyperparameters}.

\paragraph{Optimization}
Although only our \Nav{} environment involves multi-step episodes, using a full reinforcement learning algorithm across all environments benefits comparability and extensibility in future work.
Specifically, we use proximal policy optimization (PPO) \citep{schulman2017ppo} paired with Adam \citep{kingma2015adam} to optimize the neural networks.
PPO is widely used RL algorithm which selected primarily for its stability (e.g., training almost always converges, minimal hyperparameter tuning); attempts to train with ``vanilla'' advantage actor critic did not consistently converge.
We use the PPO implementation of Stable Baselines 3 (MIT license) built on PyTorch (BSD license) \citep{stable-baselines3,pytorch}.

\begin{algorithm}[t]
    \begin{lstlisting}]
n_updates: int >= 0
buffer_size: int > 0

for _ in range(n_updates):  # outer loop
  rollout_buffer = []
  for _ in range(buffer_size):  # inner loop
      episode = run_episode(model, environment)
      rollout_buffer.append(episode)
  update_parameters(model, rollout_buffer)
    \end{lstlisting}
    \caption{PPO pseudocode}\label{alg:ppo}
\end{algorithm}

One relevant characteristic of PPO and similar algorithms is that in their training they contain an inner and outer loop analogous to \theProcess{} (Algorithm~\ref{alg:the-process}); this is illustrated in Algorithm~\ref{alg:ppo}.
The (main) outer loop consists of two steps: the inner loop which populates a rollout buffer with ``experience'' from the environment and the updating of parameters based on that buffer.
What is important to note is that the buffer is populated with data from the same model parameters, and it is not until after this that model parameters change.

% It is important to note that the model makes simplifying assumptions to the ELSs which is intended to model.
% For example, while the ELSs use a Gumbel-Softmax layer to sample from the lexicon, \theProcess{} simply uses a multinomial distribution
% This is done so as to keep \theProcess{} simple enough to conceptually grasp.

\subsection{Hypothesis}
Here we state the hypothesis used to evaluate \theProcess{}.
The sign of hyperparameter-entropy correlation observed in \theProcess{} will be the same as what we observe for a corresponding hyperparameter in the ELSs.
We can state this more formally as: for each pair of corresponding hyperparameters $(h, h')$ in \theProcess{} and an ELS respectively,
\begin{align}
    \text{sgn}(\text{corr}(D)) &= \text{sgn}(\text{corr}(D')) \\
    D &= \left\{ (x, H(\bs y)) \mid x \in X_h,\, \bs y \sim \theProcess{}_{h=x} \right\} \\
    D' &= \left\{ (x, H(\bs y)) \mid x \in X_{h'},\, \bs y \sim \text{ELS}_{h'=x} \right\} \\
   H(\bs y) &= -\sum_{i=1}^S y_i \log_2 y_i
\end{align}
where
    $\text{corr}(\cdot)$ is the Kendall rank correlation coefficient ($\tau$) \citep{kendalltau},
    $\theProcess{}_{h=x}$ is the distribution over frequency vectors yielded by the model for hyperparameter $h$ set to $x$ (assume likewise for $\text{ELS}_{h'=x}$),
    $H$ is Shannon entropy, and
    $X_h$ is the set of experimental values for hyperparameter $h$.
A ``sample'' from an ELS consists of training the agents in the environment, and estimating word frequencies by collecting the sender's messages over a random sample of inputs.
Accordingly, our null hypothesis is that \theProcess{} does not meaningfully correspond to the ELSs, and thus the signs of correlation would be expected to match with a probability $0.5$.

We intentionally formulate our hypothesis at this level of granularity:
    equality of direction (sign) of correlation rather stronger claims
    such as raw correlation: $|\text{corr}(D) - \text{corr}(D')| < \epsilon$
    or mean squared error: $1/|X| \cdot \sum_{x\in X} (D(x) - D'(x))^2$.
We select this level of direction of correlation for a few reasons.
The level of simplicity of \theProcess{} compared to the ELSs means that the unaccounted for factors would make supporting stronger hypotheses too difficult;
    furthermore, even if the hypothesis were defended, it would be less widely applicable for the same reasons.
Additionally, the current literature tends to speak of the general principles of emergent language at the level of ``relationships'' and ``effects'' rather than exact numeric approximations \citep{kharitonov2020entropy,resnick_capacity_2020}.

\begin{table}
    \centering
    \caption{Corresponding hyperparameters in the ELSs and \theProcess{}.}%
    \label{tab:correspondence}
    \begin{tabular}{lc}
        \toprule
        ELS & \theProcess{} \\
        \midrule
        Time steps & $N$ \\     
        Lexicon size & $S$ \\     
        Learning rate & $\alpha$ \\     
        Buffer size & $\beta$ \\     
        Temperature & $\beta$ \\     
        \bottomrule
    \end{tabular}
\end{table}

\paragraph{Corresponding Hyperparameters}
A key component of the hypothesis is the correspondence of hyperparameters of the ELSs with those of \theProcess{}.
These correspondences are the foundation for applying reasoning about \theProcess{} to the ELSs; accordingly, they also determine how the model will be empirically tested.
We present five pairs of corresponding environment-agnostic hyperparameters in Table~\ref{tab:correspondence}.
Although environment-specific hyperparameters can easily correspond with those of \theProcess{} we chose the agnostic for ease of experimentation and comparison.

To identify these correspondences, it is important to understand the intuitive similarities between the ELSs and \theProcess{}.
Firstly, the weights of \theProcess{} correspond the learned likelihood with which a given bottleneck unit is used in the ELS\@; in turn, both of these correspond to the frequency with which a word is used in a language.
Each iteration of \theProcess{}'s outer loop is analogous to a whole cycle in the ELS of simulating episodes in the environment, receiving the rewards, and performing gradient descent with respect to the rewards (compare Algorithms~\ref{alg:the-process} and~\ref{alg:ppo}).

Based on this analogy, we can explain the corresponding hyperparameters as follows.
$N$ corresponds the number of parameter updates taken throughout the course of training the ELS\@ (i.e., the outer loop of PPO).
$S$ corresponds the size of the bottleneck layer in the ELS\@.
$\alpha$ corresponds to the learning rate (i.e., magnitude of parameter updates) in the ELS\@.
The ELS has two analogs of $\beta$.
First, $\beta$ corresponds to the rollout buffer size of PPO because both control the number of iterations of the inner loop of training where episodes are collected before updating the weights.
Second, $\beta$, more generally, control how smooth the updates to \theProcess{}'s weights are which makes it analogous to the temperature of the Gumbel-Softmax distribution in the ELS\@ since a higher temperature results in smoother updates to the bottleneck's parameters.

% \cmg{discuss analogy of learned words are used more and used words are learned more?}
% \cmg{talk about type-token distinction?}

\section{Experiments}\label{sec:experiments}

Our experiments consist of comparing the correlation between the hyperparameters of \theProcess{} and the ELSs and the Shannon entropy of lexicon at the end of training.
The entropy for the ELSs is calculated based on the bottleneck unit (word) frequencies gathered by sampling from the sender's input distribution.
To gather data for \theProcess{}, we run a Rust implementation of a sampling algorithm.
Each experiment consists of a logarithmic sweep of a hyperparameter plotted against the entropy yielded by those hyperparameters (see Appendix~\ref{sec:hyperparameters} for details).

\begin{figure}
    \newcommand\mksubfig[1]{%
        \begin{subfigure}[t]{0.17\linewidth}
            \centering
            \includegraphics[width=\linewidth]{assets/plots/#1}
        \end{subfigure}
    }
    \newcommand\mkhead[1]{%
        \begin{subfigure}[t]{0.17\linewidth}
            \centering
            #1
        \end{subfigure}
    }

    \hfill
    \mkhead{Time Steps}
    \mkhead{Lexicon Size}
    \mkhead{Learning Rate}
    \mkhead{Buffer Size}
    \mkhead{Temperature}
    \vspace{1ex}

    \rotatebox{60}{\hspace{1.5em}\theProcess}
    \hfill
    \mksubfig{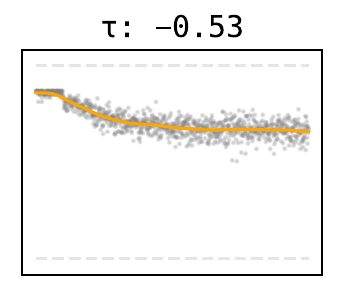}
    \mksubfig{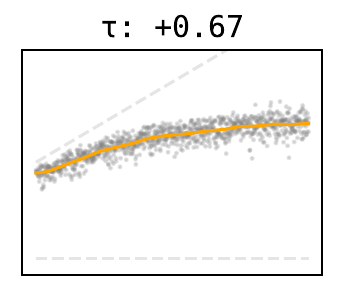}
    \mksubfig{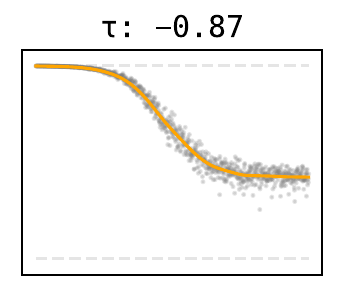}
    \mksubfig{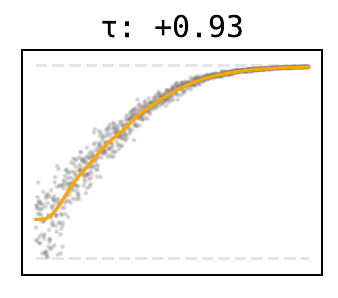}
    \mksubfig{model-beta}

    \rotatebox{60}{\hspace{1.5em}\Nodyn}
    \hfill
    \mksubfig{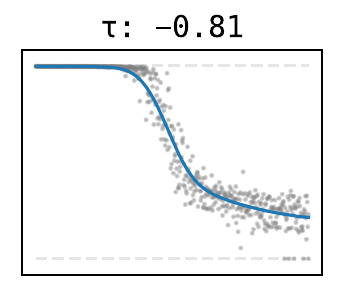}
    \mksubfig{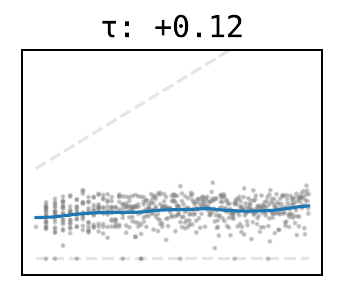}
    \mksubfig{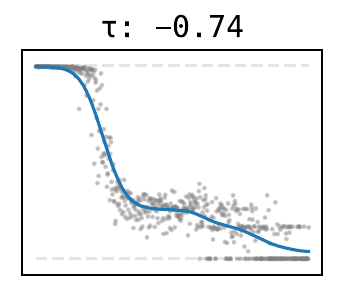}
    \mksubfig{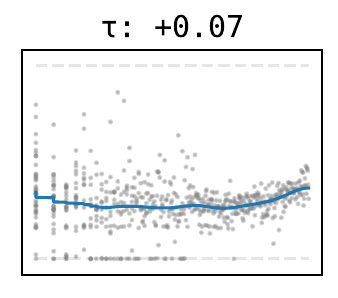}
    \mksubfig{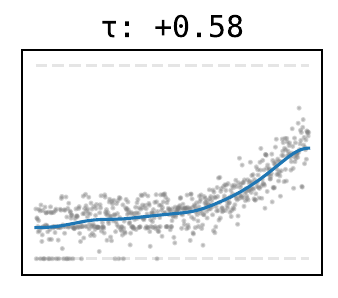}

    \rotatebox{60}{\hspace{1.5em}\Recon}
    \hfill
    \mksubfig{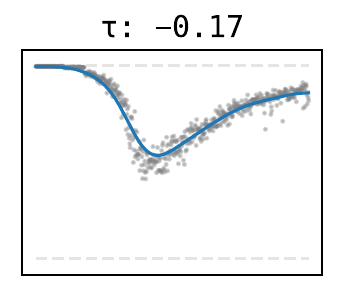}
    \mksubfig{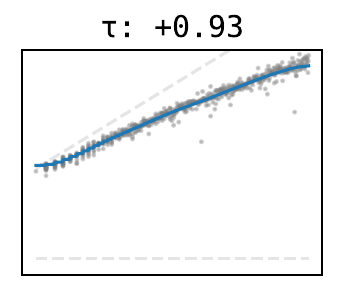}
    \mksubfig{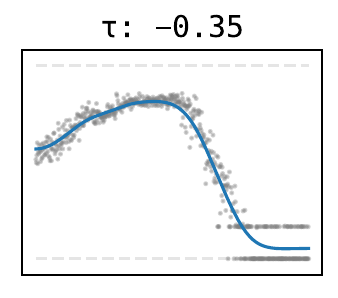}
    \mksubfig{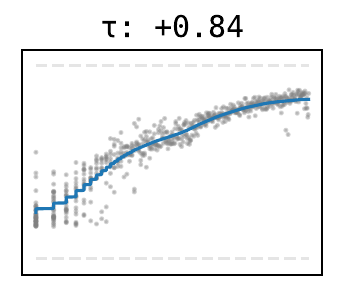}
    \mksubfig{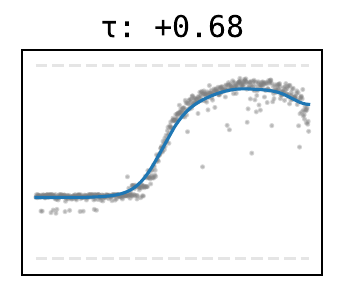}

    \rotatebox{60}{\hspace{1.5em}\Sig}
    \hfill
    \mksubfig{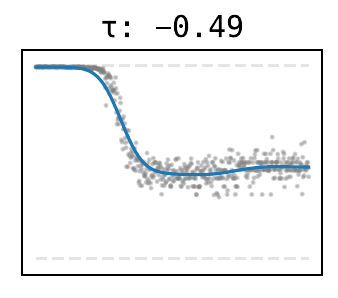}
    \mksubfig{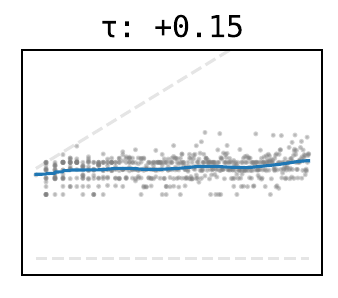}
    \mksubfig{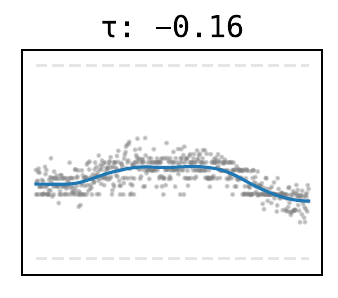}
    \mksubfig{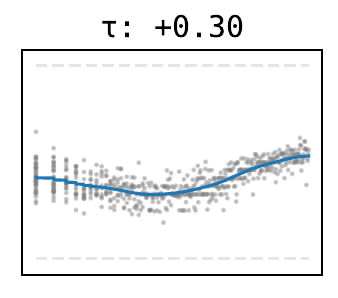}
    \mksubfig{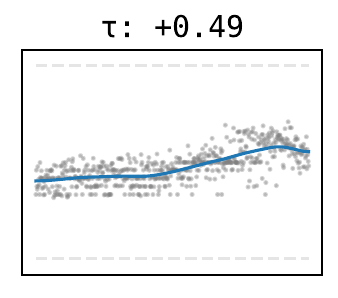}

    \rotatebox{60}{\hspace{1.5em}\Nav}
    \hfill
    \mksubfig{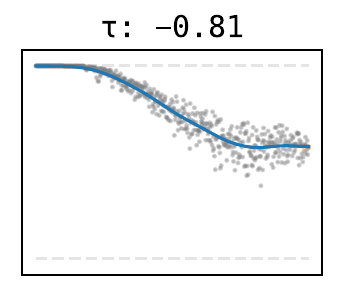}
    \mksubfig{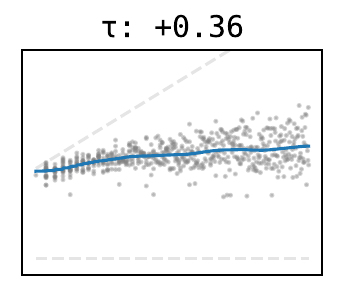}
    \mksubfig{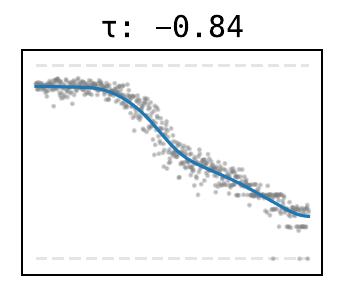}
    \mksubfig{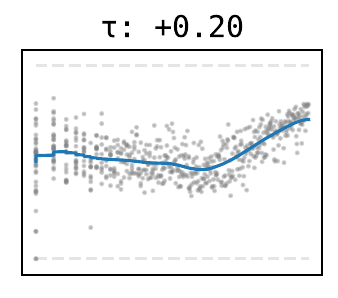}
    \mksubfig{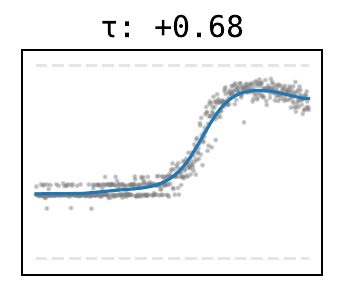}

    \caption{%
        Plots of hyperparameters ($x$-axis, log scale) vs.\@ entropy ($y$-axis) .
        Each row corresponds to a particular environment.
        Each column corresponds to a particular hyperparameter.
        All $y$-axes are on the same scale with the dashed lines representing min/max entropy.
        The points are individual runs and the lines are a Gaussian convolution of the points.
    }\label{fig:plots}
\end{figure}

\begin{table}
    \caption{
        Kendall's $\tau$'s for various configurations.
        All values have a significance of $p\le0.01$.
        % ``---'' denotes no statistically significant correlation.
    }\label{tab:corr}

    \centering
    \begin{tabular}{lrrrrr}
    \toprule
    Environment & Time Steps  & Lexicon Size   & Learning Rate  & Buffer Size  & Temperature  \\
    \midrule
    \theProcess{}   & $-0.53$ & $+0.67$ & $-0.87$ & $+0.93$ & $+0.93$ \\
    \Nodyn{}        & $-0.81$ & $+0.12$ & $-0.74$ & $+0.07$ & $+0.58$ \\
    \Recon{}        & $-0.17$ & $+0.93$ & $-0.35$ & $+0.84$ & $+0.68$ \\
    \Sig{}          & $-0.49$ & $+0.15$ & $-0.16$ & $+0.30$ & $+0.49$ \\
    \Nav{}          & $-0.81$ & $+0.36$ & $-0.84$ & $+0.20$ & $+0.68$ \\
    \bottomrule
    \end{tabular}
\end{table}

Each point in the resulting scatter plots corresponds to an independent run of the model or ELS with the hyperparameter on the $x$-axis and entropy on the $y$-axis.
The plots also include a Gaussian convolution of the data points (the solid line) to better illustrate the general trend of the data.
The plots are presented in Figure~\ref{fig:plots} with the rank correlation coefficients in Table~\ref{tab:corr}.

\section{Discussion}\label{sec:discussion}

\subsection{Model evaluation}\label{sec:model-eval}
Looking at the signs of correlations shows that \theProcess{} makes the correct prediction $20$ out of $20$ times.
Given a simple one-sided binomial test, the empirical data rejects the null hypothesis at $p<0.001$.
Although this number drops to $15$ out of $20$ if we require $|\tau| \ge 0.2$, the binomial test rejects the null hypothesis with $p=0.02$ for this stronger hypothesis.

Though the directions of correlations predicted by \theProcess{} are correct, looking at the plots show that ELSs do not always demonstrate the monotonicity predicted by the model.
This is especially evident in \emph{Time Steps} for \Recon{}:
    moving left-to-right, the plot follows a similar path to the other environment and \theProcess{} at first but then diverges halfway through with increasing entropy.
A possible explanation of this is that \Recon{} allows learning new, useful words more easily than \Sig{} or \Nav{}, meaning that additional training can lead to further improvement.
The conclusion we draw from these plots is that \theProcess{} correctly predicts a sort of baseline correlation between the hyperparameters and entropy.
Other works, \citet{kharitonov2020entropy,chaabouni2021color} for example, find similar correlations between entropy and bottleneck temperature.
Nevertheless, this correlation can be overridden by the specifics of the environment.
% In this way, the model is intended to be a starting point both for reasoning about yet unstudied environments and configurations and for building more detailed and faithful models of specific environments.

\subsection{Environment variability}
% \cmg{talk about confounding effects like world size for \Nav?}

When looking beyond just the direction of correlation at the slopes and shapes of the curves, the four ELSs all present unique set of relationships between entropy their hyperparameters.
This implies that none of these environments are reducible to each other, that is, we cannot make observations about one environment and automatically assume they apply to other environments.
Certainly this makes an researcher's task harder as learning general principles would not be possible from a single environment.
Furthermore, there is a sensitivity to hyperparameters \emph{within} a given environment, which would imply that discovering general principles within single environment could not be done with just a single set of hyperparameters.

Although this diversity in behavior makes modeling it more difficult, it also shows the importance of precision we get from a mathematical model.
% Specifically, models help articulate precisely the generalizations diverse environments.
For example, say \Recon{} has not been empirically tested and we wanted to predict the lexicon size-entropy relationship in \Recon{}.
It is the case that we could simply observe the positive correlations in the other environments and predict the same \Recon{}, but we could easily over-extrapolate and predict a relatively shallow slope when \Recon{}'s slope is relatively steep.
What this paper's model, hypothesis, and evaluation offer in this situation is not a more detailed prediction but a  ``prepackaged'' prediction which is precisely stated and supported by data.
% Furthermore, since \theProcess{} is the mathematical model from which the hypothesis is derived, we could in turn make further predictions based on the model beyond what was explicitly tested (e.g., a new ELS hyperparameter analogous to a hyperparameter of \theProcess{}).

\subsection{Applications to future work}
There are two primary ways in which \theProcess{} can be applied in future research.
First, the model can be applied to and tested against further phenomena in emergent language (i.e., it is \emph{extensible}).
The fact that it is formulated mathematically means that it does not just predict correlations but \emph{mechanisms} which account for the correlations.
For example, \theProcess{}'s $\beta$ hyperparameter was designed to account for \emph{Buffer Size} and the \emph{Temperature} experiment was conducted after the fact.
The fact that \theProcess{} describes both \emph{Buffer Size} and \emph{Temperature} with the same hyperparameter suggests that similar mechanisms account for their positive correlations with entropy.
This statement about similar mechanisms, on the other hand, is not present set of one-off hypotheses about hyperparameter-entropy correlations derived from intuition.
Second, \theProcess{} and accompanying experiments provide an easy way for future research to discover confounding factors in their experiments.
For example, an experiment might show that entropy decreases as rewards are scaled up, yet \theProcess{} would suggest that this might be equivalent to simply increasing the learning rate rather than being its own unique cause of the effect on entropy.
% \cmg{I could get this, though, from a collection of many experiments; what a model offers is an explanation of \emph{how} it works---this should be mentioned either way---but is there an example of where this is clearly the case with \theProcess{}?}

\subsection{Methodological difficulties}\label{sec:methodological}
The greatest challenge in the methodology of this work is not the formulation of the model but rather evaluating the quality of the model.
In part, this is on account of a lack of established baseline model---comparative analysis (``which is better?'') is significantly easier than absolute analysis (``how good is this?'') yet requires an adequate baseline to compare against.
But more significantly, the granularity of experimentation is a design decision with no obvious answer.

For example, merely comparing the signs of rank correlations is very coarse-grained as it makes minimal assumptions about the data (e.g., linearity, absence of outliers) and captures very little information about the data.
Naturally, it is easier to apply such an analysis, and as mentioned before, researcher typically phrase hypotheses in terms of such correlations, but it can only offer minimal support for applicability of the model to the actual system.
On the other hand, evaluating the model's ability to predict exact behavior of the system (e.g., measuring mean squared error of the model's predictions) can establish a more precise link between model and system but might miss more general but important similarities.
For example, \emph{Lexicon Size} for \theProcess{} and \Nav{} might show similar trends, but be different by a constant, yielding a high mean squared error.

A subtle but significant methodological difficulty is the selection of hyperparameters.
In \Recon's \emph{Time Steps} plot, it is easy to see that changing the range of hyperparameters could easily yield either a positive or a negative correlation when in reality there are both.
To a certain extent, this can be resolved be choosing a ``reasonable'' range of hyperparameters based on values are typically, but this is of little help to selection of \theProcess{}'s hyperparameters as there is no ``typical usage.''
For example, \theProcess{} for $\beta=1$ and $\beta=100$ yield significantly different distributions, but there is no obvious \emph{a priori} reason to say that one value of $\beta$ should be preferred over the other for comparing to the ELSs.
Although additional hyperparameters increase the range of phenomena which the model can account for, the additional degrees of freedom can weaken the model's predictions by introducing confounding variables (cf.\@ overparameterization).

% Naturally, many generalizations of this model are possible: applying it to new environments, accounting for multi-word utterances, using different optimization schemata, and incorporating reward dynamics.
One of the primary contributions of this work is to serve as a case study and example of working with explicitly defined models in studying deep learning-based emergent language.
Thus, this paper is starting point for future work to improve upon.
One of the most important improvements would be finding a more rigorous way to select ``reasonable'' experimental hyperparameters.
Additionally, it would be better to develop the hypothesis and experimental in full before performing any evaluation; the process was somewhat iterative in this paper.

\section{Conclusion}\label{sec:conclusion}

We have presented \theProcess{} as a mathematical model of lexicon entropy in deep learning-based emergent language systems and demonstrated that, at the level of correlations, it accurately predicts the behavior of our emergent language environments.
Opting for a mathematical model possesses the benefits of having a clear interpretation, making testable predictions, and being reused for new predictions in future studies.
Although the model's hypothesis was testable, the process is not free from non-trivial design decisions which affect the quality of evaluation.
Nevertheless, this paper serves as starting point and example of how more rigorous models can be applied to the study of emergent language.

\begin{ack}
This material is based on research sponsored in part by the Air Force Research Laboratory under agreement number FA8750-19-2-0200. The U.S.  Government is authorized to reproduce and distribute reprints for Governmental purposes notwithstanding any copyright notation thereon. The views and conclusions contained herein are those of the authors and should not be interpreted as necessarily representing the official policies or endorsements, either expressed or implied, of the Air Force Research Laboratory or the U.S. Government.
\end{ack}

\bibliography{main}
\bibliographystyle{plainnat}

\appendix

\section{Emergent language system illustration}\label{sec:illustration}

\begin{figure}[H]
    \centering
    \def\figheight{20ex}
    \begin{subfigure}[t]{0.40\linewidth}
        \centering
        \includegraphics[height=\figheight]{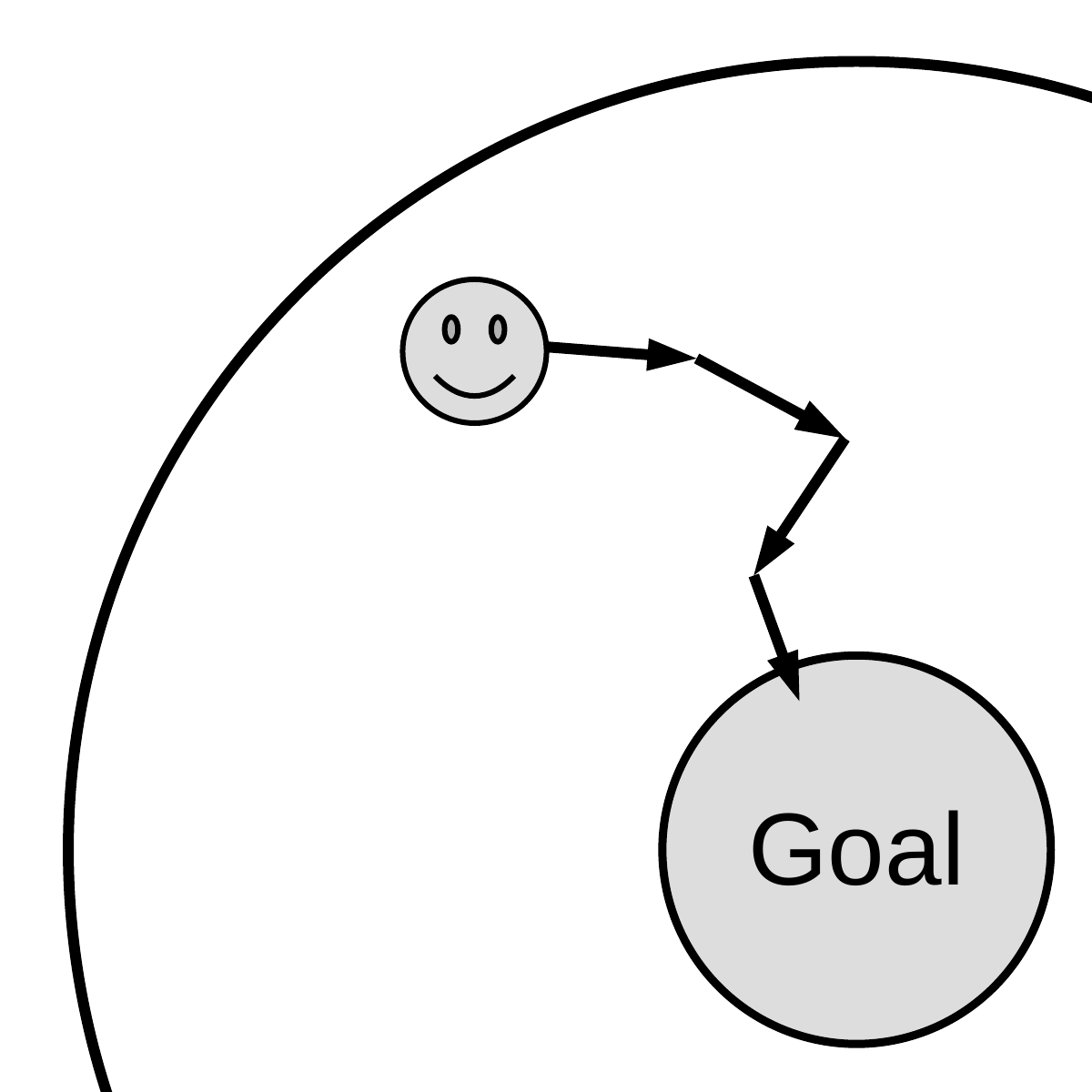}
        \caption{The receiver (pictured) is rewarded for moving towards the goal region at the center in the \Nav{} environment.}\label{fig:env}
    \end{subfigure}
    \hfill
    \begin{subfigure}[t]{0.40\linewidth}
        \centering
        \includegraphics[height=\figheight]{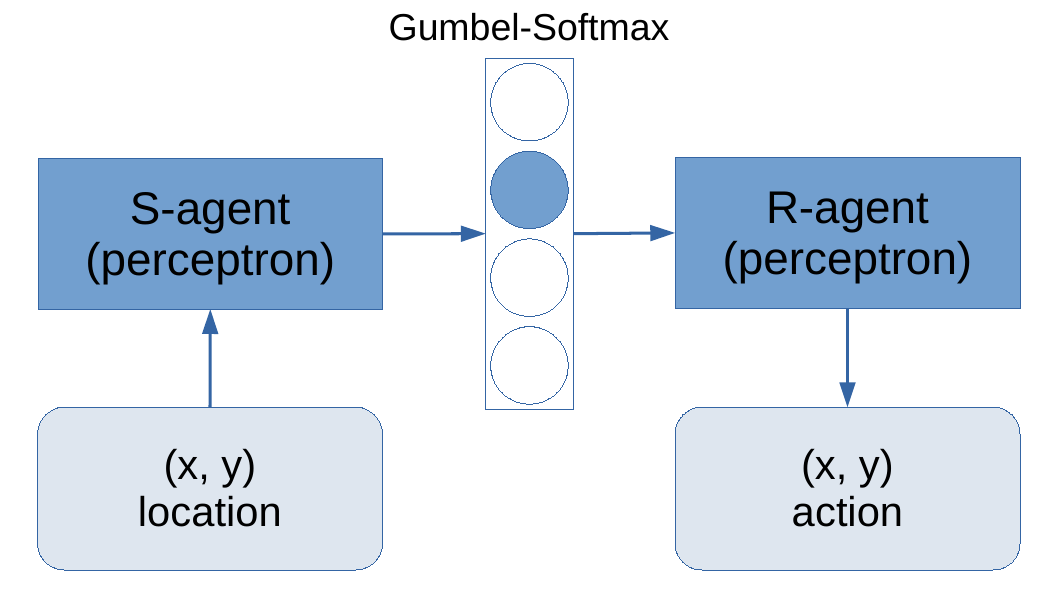}
        \caption{The agent architecture for \Nav{}.}\label{fig:arch}
    \end{subfigure}
    \caption{}
\end{figure}

\section{Experiment parameters}\label{sec:hyperparameters}

Each experiment uses a logarithmic sweep across hyperparameters; the sweep is defined by Equation~\ref{eq:log-sweep}, where $x$ and $y$ are the inclusive upper and lower bounds respectively and $n$ is the number steps to divide the interval into.
The floor function is applied if the elements must be integers.
\begin{equation}
    \text{LS}(x,y,n)=
    \left\{x\cdot\left(\frac{y}{x}\right)^{\frac{i}{n-1}} \,\middle|\, i \in \{0, 1, \dots, n-1\}\right\}
    \label{eq:log-sweep}
\end{equation}

\begin{table}[H]
    \centering
    \begin{tabular}{lrrrr}
        \toprule
        Hyperparameter & Default & Low & High & Steps \\
        \midrule
        $N$         & $10^3$ & $10^{0}$ & $10^{3}$ & $1000$ \\
        $S$         & $2^6$ & $2^{3}$ & $2^{8}$ & $1000$ \\
        $\alpha$    & $1$ & $10^{-3}$ & $10^{3}$ & $1000$ \\
        $\beta$     & $8$ & $10^{0}$ & $10^{3}$ & $1000$ \\
        \bottomrule
    \end{tabular}
    \caption{%
        Hyperparameters for the empirical evaluation of \theProcess{}.
        ``Low'' and ``High'' refer to the logarithmic sweep used for that experiment; default values used for all other experiments.
    }\label{tab:filex-hps}

\end{table}

\begin{table}[H]
    \centering
    \begin{tabular}{lrrrr}
        \toprule
        Hyperparameter & Default & Low & High & Steps \\
        \midrule
        Time steps      & $2\cdot10^5$ & $10^{2}$ & $10^{6}$ & $600$ \\
        Bottleneck size & $2^6$ & $2^{3}$ & $2^{8}$ & $600$ \\
        Learning rate   & $3\cdot10^{-3}$ & $10^{-4}$ & $10^{-1}$ & $600$ \\
        Buffer size     & $2^8$ & $2^{3}$ & $2^{10}$ & $600$ \\
        Temperature     & $1.5$ & $10^{-1}$ & $10^{1}$ & $600$ \\
        \bottomrule
    \end{tabular}
    \caption{%
        Hyperparameters for the empirical evaluation of \theProcess{}.
        ``Low'' and ``High'' refer to the logarithmic sweep used for that experiment; default values used for all other experiments.
        Please see code for further details and default values.
    }\label{tab:els-hps}
\end{table}

\end{document}